\documentclass{article}

\usepackage[nonatbib,preprint]{neurips_2020}

\usepackage[utf8]{inputenc} 
\usepackage[T1]{fontenc}    
\usepackage{hyperref}       
\usepackage{url}            
\usepackage{booktabs}       
\usepackage{amsfonts}       
\usepackage{nicefrac}       
\usepackage{microtype}      
\usepackage{amsmath,amsthm,amssymb}
\usepackage{graphicx}
\usepackage{verbatim}
\usepackage{float}
\usepackage{color}
\usepackage{dirtytalk}
\usepackage{bbm}
\usepackage{enumitem}
\usepackage[ruled,linesnumbered]{algorithm2e}
\usepackage{xcolor}
\usepackage{multirow}
\usepackage{pgfplots}
\usepackage{threeparttable}
\usepackage{caption}
\usepackage{subcaption}
\usepackage{wrapfig}
\usepackage{balance}  
\usepackage{tikz}
\usepackage{xr}

\newtheorem{example}{Example}[section]
\newtheorem{corollary}{Corollary}[section]
\newtheorem{definition}{Definition}

\newcommand{\mc}[1]{\mathcal{ #1}}
\newcommand{\mb}[1]{\mathbf{ #1}}

\DeclareMathOperator{\argmax}{argmax}

\author{%
  Alireza~Heidari \\
  Department of Computer Science\\
  University of Waterloo\\
  \texttt{a5heidar@uwaterloo.ca} \\
   \And
   George~Michalopoulos\\
   Department of Computer Science\\
  University of Waterloo\\
  \texttt{gmichalo@uwaterloo.ca} \\
   \And
   Shrinu~Kushagra \\
  Department of Computer Science\\
  University of Waterloo\\
  \texttt{skushagr@uwaterloo.ca} \\
   \And
     Ihab F.~Ilyas \\
  Department of Computer Science\\
  University of Waterloo\\
  \texttt{ilyas@uwaterloo.ca} \\
     \And
     Theodoros~Rekatsinas \\
  Department of Computer Science\\
  University of Wisconsin-Madison\\
  \texttt{thodrek@cs.wisc.edu} \\
}

\title{Record fusion: A learning approach}
\begin{document}

\maketitle

\begin{abstract} 
Record fusion is the task of aggregating multiple records that correspond to the same real-world entity in a database. We can view record fusion as a machine learning problem where the goal is to predict the \say{correct} value for each attribute for each entity.  Given a database, we use a combination of attribute-level, record-level, and database-level signals to construct a feature vector for each cell (or (row, col)) of that database. We use this feature vector alongwith the ground-truth information to learn a classifier for each of the attributes of the database. 

Our learning algorithm uses a novel stagewise additive model. At each stage, we construct a new feature vector by combining a part of the original feature vector with features computed by the predictions from the previous stage. We then learn a softmax classifier over the new feature space. This greedy stagewise approach can be viewed as a deep model where at each stage, we are adding more complicated non-linear transformations of the original feature vector. We show that our approach fuses records with an average precision of $\sim$98\% when source information of records is available, and $\sim$94\% without source information across a diverse array of real-world datasets. We compare our approach to a comprehensive collection of data fusion and entity consolidation methods considered in the literature. We show that our approach can achieve an average precision improvement of $\sim 20 \%/\sim 45 \%$ with/without source information respectively. 
\end{abstract}

\vspace{-5pt}
\section{Introduction}
\vspace{-8pt}
Modern enterprises generate a huge amount of data. This data is then used to make many business-critical and often mission-critical decisions. However ensuring consistency and correctness of the database records is a challenging task.  With passage of time, errors can creep into these datasets, some being out-of-date, inexact, or incorrect. Besides human error, a major source of errors is the following. Often that database is accumulated from multiple different sources. This can lead to the same physical entity having multiple different records in the database. This severely affects the quality of the downstream analytics. The increasing demand to ingest and to acquire large number of heterogeneous data source via inexpensive connectors has been the driving force behind many studies in the field of entity consolidation and data integration \cite{bruggemann1998one,dong2009truth,dong2018data,dong2012less,heidari2019approximate,li2014confidence,li2012truth,yin2008truth}.  

\begin{figure}
  \centering
  \includegraphics[width=0.7\columnwidth]{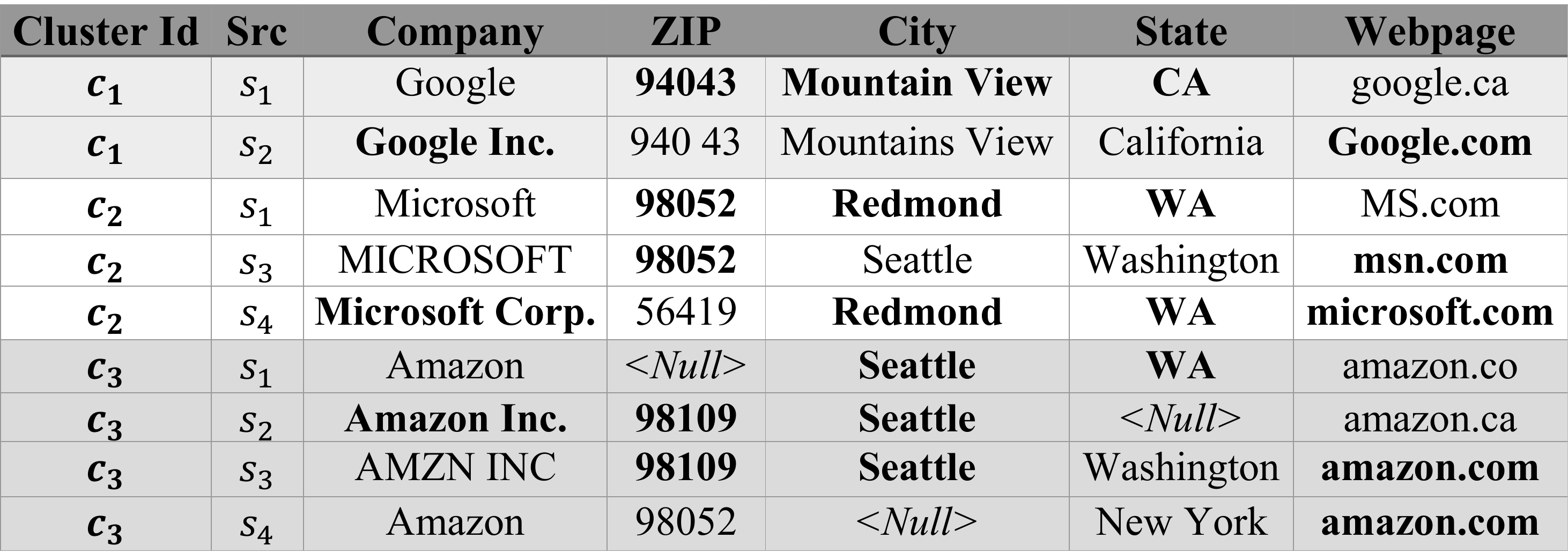}
  \caption{An example dataset where the records have been gathered from multiple sources. There are three entities or clusters. The correct value for each attribute is shown in \textbf{Bold}. Each row of the database makes a claim about the actual value for this real-word entity. However, pieces of information that are gathered from different sources can be conflicting. For instance, the first source of cluster $c_3$ claims that the Amazon headquarters is located in state "WA", while, the last source claims that the state of the entity is "New York". In the record fusion problem, we want to resolve such conflicts and obtain the correct values for the attributes of each real-world entity.}
  \label{fig:exampletab}
  \vspace{-10pt}
\end{figure}

A common approach to solve the data integrity problem in the aforementioned paragraph is the following. Given a dataset, first detect all the records which correspond to the same real-world entity. This problem is referred to as data dedupliction or record deduplication \cite{chu2016distributed,deng2017unsupervised,elmagarmid2006duplicate,getoor2012entity}. Once the entities are detected, the next challenge is to merge the records of the same entity (possibly conflicting) into one single record. This is referred to as \textit{Record Fusion} and is the focus of this paper. The problem of merging records corresponding to the same entity has been extensively studied in the database community and also referred to as \textit{the golden record problem} or the \textit{data fusion problem} \cite{dong2009truth,dong2012less,li2012truth,yin2008truth}. Fig. \ref{fig:exampletab} gives a small toy example of the record fusion problem. We will refer back to this during the paper as a suggestive example for some of our algorithms.

Previous approaches to tackle the record fusion problem make use of source information. In such situations, for every record, on top of the cluster identities the source that generated the tuple is also known. These works then build models to estimate the \say{trustworthiness} of the different sources \cite{dong2013compact,wu2011framework,yin2008truth,yin2011semi,dong2009truth}.  For example, in some instances, a limited ground truth is used for calculating an initial estimation of the sources \say{trust score}\cite{dong2009integrating,galland2010corroborating}. However, reliably estimating the trustworthiness of a source is non-trivial. Furthermore, it is unclear that estimating trustworthiness of a source is the only signal that determines how the records should be merged. In some scenarios, all sources might be equally good/bad and theses source-dependant methods reduce to simple majority voting.

 Moreover, in many real-world scenarios, we may not have explicit source information altogether. In such scenarios, current approaches either use techniques like heuristic aggregation rules, majority vote, or involve humans to resolve conflicts via learned transformations \cite{deng2017unsupervised,heidari2019holodetect}. We show experimentally that na\"ive methods like choosing the value provided by the majority vote of sources often lead to inaccurate and unreliable results (see Section \ref{exp:e2e}). Human-in-the-loop approaches have multiple challenges, including (1)~they can be prohibitively expensive and time-consuming depending on the number of clusters, and the difficulty of resolving conflicts; and (2)~they often assume users do not make mistakes, or they need to involve multiple voters introducing new types of conflicts that needs additional resolution mechanisms.

In this paper, we show how to use a principled machine learning approach to solve the record fusion problem taking into account all available signals (statistical properties and constraints). We propose a learning framework for record fusion based on weak supervision \cite{hoffmann2011knowledge,ratner2017snorkel}; our framework automatically fuses records by leveraging all related information, i.e., integrity constraints, and quantitative statistics, and source information if available. We show that our approach is able to fuse records when source information is unavailable with an average precision $\sim$94\%; an improvement of  $\sim 45\%$ over previous approaches, also obtaining an average precision of $\sim$98\%; an improvement of over $\sim 20\%$ from previous approaches, when source information is available.

 \vspace{-4pt}
 \subsection{Technical challenges and contributions}
 \vspace{-7pt}
 
 Our ML-approach addresses multiple technical challenges with concrete contributions. We highlight these challenges and solutions in the following sections.
	
	 \vspace{-8pt}
	\subsubsection*{Feature representation}
	 \vspace{-8pt}
	 A characteristic of record fusion is homogeneity within an entity and heterogeneity across entities. Hence, any feature design methodology needs to take this into account. As a concrete example, consider the database in Fig. \ref{fig:exampletab} and lets focus on the first column `Company Name'. For each entity, the values for the first column are similar (`Google', `Google Inc.' ) but are very different when compared against that column values across entities, for example `Amzn', `Microsoft' etc. Hence, na\"ive representations like one-hot encoded strings or even \textit{word2vec} representations are not likely to be able to uncover the relationship that governs which value is correct across that particular entity.
	 
	  To address this issue, we design representation models that capture various characteristics of the data that providing a rich input to a model. Attribute-level features which capture the distributions governing the values and format of the cell attribute. Tuple-level features capture the joint distribution of different attributes and perform weak predictions for possible values of each cell of final dataset. Dataset-level features capture a distribution that governs the compatibility of tuples and values in the dataset $D$. Section \ref{sec:featuredesign} provides details of  each of these steps. 
	
	\vspace{-8pt}
    \subsection*{Model construction and Training Data Generation}
    \vspace{-8pt}
    	Another property of the record fusion domain is the heterogeneity across different attributes. That is each, column of the database is very different from other columns and sometimes has its own unique format (e.g., Addresses, Zip codes, Webpages see Fig. \ref{fig:exampletab}). Representing all these different features in a single domain is a challenge. A learning algorithm which tries to predict the correct value of cells over the range of all possible values with heterogeneous formats is deemed to fail or require prohibitively large number of data points. To tackle this problem, we propose to learn a different classifier for each attribute. Hence, our learning framework outputs $c$ different classifiers where $c$ is the number of columns in the dataset. 
	    
	    Once the features are constructed, we use a stage-wise additive model to learn a classifier. First, we learn a softmax classifier using our original featurized dataset. While it is possible to use this classifier to make predictions, we construct deeper models using non-linear features using a greedy stage-wise approach. More concretely, we use the predictions from the previous stage to construct a new feature set for current stage which comprises of the original features and also the `dynamic features' from the previous step. We again learn a logistic regression classifier using these sets of features. We repeat this process for a fixed number of steps. 
    
        While the approach described in the previous items is sufficient for a comprehensive treatment of the record fusion problem. We go one step further and also \textbf{add data augmentation} amidst this mix so that our models are more robust. We discuss more about it in Section \ref{sec:learningalgorithm} and appendix. Adding a data-augmentation at $10\%$ seems to be the most effective in improving classifier accuracy. 
	    
	    Finally, we evaluate our framework on multiple real-world datasets, where we demonstrate its ability to determine the correct values for real-world entities, and we show that probabilistic inferences with sufficient training data are a valid modelling tool for the record fusion problem (Section~\ref{sec:exp}). 
\vspace{-5pt}
\section{Preliminaries}
\label{sec:GR}
\vspace{-8pt}
A relational database is a set of records $D = \{d_1, \ldots, d_n\}$ such that each record (or tuple) $d_i = (d_{i1}, \ldots, d_{ic})$ has $c$ attributes or columns. For all $i$, let $d_{ij}\in R_j$. We say that the database $D$ has a schema $S = \{R_1, \ldots, R_c\}$. Each $d_{ij}$ represents a cell in the database $D$.  

A clustering $\mc C$ of the relational dataset $D$ is a partition of its records into disjoint sets. That is $\mc C = \{E_1, \ldots, E_p\}\}$ where $E_k \cap E_{k'} = \varnothing$ and $\cup E_k = D$. We also say that the $E_1, \ldots, E_p$ are the set of $p$ entities for the dataset. For each $i$, we define $Id(i) := k$ if and only if $d_i \in E_k$. In this paper, we assume that the true cluster identities are known for all the records in the database. For each $1 \le k \le p$ and for each $1 \le j \le c$, define $E_{kj} = \{d_{ij} : d_i \in E_k\}$. That is, $E_{kj}$ represents the set of all possible values for the $k^{th}$ entity on the $j^{th}$ column. 

Let $G=\{g_1,\dots,g_p\}$ be a data set with schema $S$ that contains the correct label of clusters in $\mc C$. For each row $g_i=\{g_{i1},\dots,g_{ic}\}\in G$ that corresponds to cluster $c_i\in \mc C$ and attribute $R_j\in S$, $g_{ij}$ is a $\rho_j$-dimensional one-hot vector that represent the correct values, where $d^j = \max_i E_{i j}$. 

Besides $D$, we are provided with a labels set $G_T\subset G$. $g_{ij}$ is given for all attribute $R_j\in S$ and every Tuple $g_i\in G_T$, and also, optionally, a set of data rules as in form of denial constraints (Def. \ref{def:dc}) $\Sigma$ might provided. We define $G_U=G\backslash   G_T$ as the set of unknown labels. We are now ready to define the record fusion problem.

\begin{definition}[Record Fusion]
\label{def1}
\noindent Let $D$ be a relational database with schema $S$. Let $\mc C$ be the set of entities of the database, and $E_{kj}$ be the set of all possible values of the $k^{th}$ entity for the $j^{th}$ attribute. The goal is for each cell in $G_U$, output a probability distribution over $E_{kj}$ that determines the correct label. 
\end{definition}

To solve the record fusion problem, we use the following strategy. From the given database, we learn classifiers $h_1, \ldots, h_c$ for each of the attributes in the database. Before we introduce our framework in more detail, it is useful to define the label dimension which will be used later.

\begin{definition}[Label dimension]
	The label dimension is the maximum number of records that correspond to a particular attribute.
	$ \rho _j := \max_{k} |E_{kj}|$
\end{definition}

Let $D_j = \{d_{ij} : 1 \le i \le n\}$. That is $D_j$ is the set of all values of $j^{th}$ column of the database. For each $d_{ij} \in D_j$, we first convert it into a feature vector $v_{ij}$. The classifier $f_j$ is then trained on the dataset $Z_j = \{(v_{ij}, l_{ij}) : 1 \le i \le n\}$ where $l_{ij}$ is an $\rho_j$-dimensional one-hot encoded vector that represents the ground truth. More precisely, let $d_{ij} \in E_{kj}$ and the ground truth value for $j^{th}$ column in the $k^{th}$ entity be given by the $q^{th}$ element in $E_{kj}$. Then $l_{ij}$ has a one in the $q^{th}$ index and zero on all the other dimensions. We will also frequently use the notation $X_j = \{v_{ij} : 1 \le i \le n\}$ to denote the features set and $y_j = \{l_{ij} : 1 \le i \le n\}$ to denote the label set. Under this notation, $Z_j = (X_j, y_j)$. For cells in $G_U$ in the first iteration, we assign majority vote as their weak labels. That is we put one in the dimension that has the maximum frequency and zero in all other dimensions, and if we have more than one maximum frequency, we select one randomly. In the next iterations, we use prediction of the model in the previous iteration.

To conclude, we construct featurized datasets $Z_1, \ldots, Z_j, \ldots, Z_c$ for each of the attributes of the database. Using these datasets, we are then able to train classifiers $h_1, \ldots, h_c$. In the next section, we describe our featurization step in detail. 
\vspace{-5pt}
\section{Feature design}
\label{sec:featuredesign}
\vspace{-8pt}
Our featurization strategy can be be broadly divided into three components; attribute-based, record-level and database-level. Attribute-level features introduce signals specific to that particular cell while the other two aim to capture more global signals. Finally, we combine the features obtained through all the three strategies into a single vector. Next, we give a more detailed description of each of these strategies. Note that formal algorithmic descriptions are included with the appendix (Section \ref{sec:features}).

\vspace{-4pt}
\subsection{Attribute-level Features}
\label{sec:attrfeat}
\vspace{-7pt}
Attribute-level features use three strategies. The first is also referred to as a \textit{format model} which aims to capture the variations governing the style or format of the values. Here, each character is replaced by a special token depending on its type. Like, all alphabets are replaced by a token `A', all numbers by token `N' etc. The original string is then represented as a vector in an $n$-gram model. In this paper, we fixed $n=2$. For example, `Google'  would be represented as `AAAAAA'  while `Google Inc.' would be represented as `AAAAAASAAAS'  and similarly for other cells. After this transformation, each string is represented as a vector which represents the count of the different $n$-grams in the string ($[5, 0, 0, 0], [7, 2, 1, 0]$). Second Strategy is Running cluster-value feature. As our model uses an iterative algorithm, this signal captures the compatibility of the attribute value in a mention with the running cluster value (after inference). For each possible value, we have a flag that indicates if it was the predicted value in an earlier iteration. In the first iteration, the current Running cluster-value feature is zero for each possible value as we do not want to give more weight to the decision of the majority vote. Character and token sequence, the third strategy, makes use of an embedding matrix $M$ which maps all values of the $j^{th}$ column into an euclidean space. Given a cell $d_{ij}$, we compute its distance to the entity average representation and include this value in the feature vector. Next, lets take a look at the record-based featurization strategy.

\vspace{-4pt}
\subsection{Record-level features}
\label{sec:recfeat}
\vspace{-7pt}
Record-level features capture the `relationship' of that particular attribute with other attributes in its record (row). We have two signals. For the first signal, we include the counts of pairs of attributes. This is also referred to as \textit{co-occurrence} counts. As an example, if the value ``New York" in attribute \textit{City} appeared more often with ``United States" of attribute ``Country", the record-level signals should reveal this effect. As we have multiple values for each attribute, we use the values that are predicted in a previous iteration for each attribute in our calculations. This feature is updated in every iteration. Finally, it should be noted that we use one co-occurrence feature per pair of attributes. \textit{Vote model} as another signal, captures how often the cell entry $d_{ij}$ occurs amongst rows within its own entity. Next, we take a look at the dataset-level featurization strategies.

\vspace{-4pt}
\subsection{Database-level Features}
\label{sec:datafeat}
\vspace{-7pt}
The first strategy that we use here is to include something called a \textit{source information}. In many record fusion applications, different rows (or records) come from different sources. Assume that there are $k$ different sources, where $k$ is known a priori. Different sources have different levels of `trust'. Hence, the knowledge of whether that cell came from a particular source is relevant to determine the `correctness' of that cell. We include source information in our features as a $k$-dimensional one-hot vector. Note that, in some cases, source information might be unavailable. In those cases, we ignore this feature and include other features.
The next information that might be available is a set of data rules in form of denial constraints \cite{livshits2020approximate}. Before, we discuss how we capture this information, lets formally define a denial constraint. 

\begin{definition}[Denial constraint]

	Given a database $D$ with $c$ attributes and schema $S$. A denial constraint is a rule with the format $\phi : \forall d_\alpha, d_\beta, d_\gamma,\dots\in D, \neg(P_1\wedge\dots\wedge P_m)$ where $P_i$ is of the form $v_1\phi v_2$ or $v_1\phi C$ with $v_1,v_2\in d_x.A$, $x\in
\{\alpha,\beta,\gamma,\dots\}$, $A\in S$, $C$ is a constant, and $\phi \in \{=,\ne,\ge,\le,<,>\}$. 
\label{def:dc}
\end{definition}

For example, the data rule $Zip \implies City$ can be presented as $\neg(d_i.Zip=d_j.Zip\wedge d_i.City\ne d_j.City)$ for $d_i,d_j\in D$. This means that any two rows in the database which have the same entry for the attribute `Zip' should also have same entry for the attribute `City'. Note that the converse need not be true. That is, there can be two rows with the same value for the column `City' but have different zip codes. Denial constraints are the most general form of rules in first order logic. For each cell $d_{ij}$, we look at the number of denial constraints that involve the $j^{th}$ column. For each of these constraints, we count the number of violations of that denial constraint assuming that the current record $d_i$ is indeed correct. We include this information in the feature vector and call it \textit{constraint violation}.

As a final strategy, we include neighbourhood-based features which makes use of the embedding matrix $M$ (used earlier in the attribute-level features too) and another embedding matrix $Q$. The matrix $Q$ is able to map an entire record (or row) into an euclidean space. For each row, we combine the embeddings obtained from $M$ and $Q$ into one joint attribute-record mapping. We compute the average of such mappings across the entity and compute the distance of the current vector from the average. 

All the strategies are described in detailed in Algs. \ref{alg:attributeFeatures}, \ref{alg:recordFeatures} and \ref{alg:databaseFeatures} in the appendix. For every cell, $d_{ij}$ the complete featurization strategy is to call the three algorithms to obtain three vectors. We then combine (or concatenate) these three vectors to obtain the final representation for that cell.

 \vspace{-5pt}
\section{Learning Algorithm}
\label{sec:learningalgorithm}
\vspace{-8pt}
Section \ref{sec:featuredesign} enables us to construct training datasets $Z_j$ for all columns in our database. Recall, that $X_j = \{v_{ij} : 1 \le i \le n \}$ is the set of all features corresponding to the $j^{th}$ column while $y_j = \{l_{ij}\}$ is the set of $\rho_j$-dimensional one-hot encoded ground truth vectors and we use $Z_j = (X_j, y_j) = \{(v_{ij}, l_{ij}) : 1 \le i \le n\}$ for training the model. Our goal is to learn a mapping $f_j$ from the set $X_j$ to the set $y_j$.

Since each $f_j$ outputs a $\rho_j$-dimensional one-hot vector, we can view this as a multi-class classification problem. While the training  phase can use some of the standard frameworks from supervised learning. The prediction phase can not use the standard approach due to the peculiarities of our problem. 

More precisely, observe that each entity $E_k$ at the column $j$ has size $\rho_j\le \rho$. However, this does not present a problem during training as a one-hot vector of dimension $|E_{kj}|$ can be trivially extended to dimension $\rho_j$. However, during the inference phase, a vector of dimension $\rho_j$ needs to be mapped back to a dimension $|E_{kj}|$ to get the predictions for that cell. We will address this problem in Section \ref{section:inferenceAlgorithm}. But before that we first describe our training algorithm. 

\begin{algorithm}
	\small
	\SetAlgoLined
	\KwIn{
		Training dataset $Z = (X, y)$ where $y_i \in \{0,1\}^\rho$. Let $I_U$ determines indices of rows in $G_U$. Let $y_T$ be the labels in $G_T$. $f$ be the function that does all featurization in Section \ref{sec:featuredesign}. $\#$iterations $T$.
	}
	\KwOut{Classifier $h$}
	
	\vspace{10pt} Define $X^{[0]} := X$.\\
	Let $h^{[0]}$ be the softmax classifier obtained by training on $(X^{[0]}, y)$\\
	\vspace{5pt}\For{$t=1$ to $T$}{
	    $y^{[t]}_U=\{y^{[t]}_i=h^{[t-1]}(X^{[t-1]}):\forall i\in I_U\}$ \\
	    $y^{[t]}=y_T\cup y^{[t]}_U$\\
		Define $X^{[t]} = f(X^{[t-1]}, y^{[t]})$\\
		Let $h^{[t]}$ be the softmax classifier obtained by training on $(X^{[t]}, y^{[t]})$
	}	
	\vspace{5pt}\textbf{return} $h^{[T]}$
	\caption{Stage-wise additive learning}
	\label{alg:il}
\end{algorithm}

Given a training set $Z = (X, y)$ such that $X \in \mb R^b$ and  $y \in \{1, \ldots, \rho\}$. A softmax classifier outputs a function $f: X \rightarrow [0, 1]^{\rho}$ according to the following rule. 
$$f(x) = softmax(W x) =: \hat y$$
where $W$ is a learned matrix of size $b \times \rho$.


Alg. \ref{alg:il} describes the training procedure for any one of the datasets $Z_j$. We repeat Alg. \ref{alg:il} $c$ times to get classifiers $h_1, \ldots, h_c$. For a particular $Z = (X, y)$, the algorithm works as follows. We first learn a softmax classifier $h^{[0]}$ over the entire dataset $X$. We then use the output $h_0(x)$  to make better prediction for unlabeled data and concatenate it training data labels $y^{[1]}$. Then we update the dynamic feature vectors, the ones that need labels to be computed. More precisely, $X^{[1]} = \{f(x_i^{[0]}, y_i^{[1]}) : x_i \in X\}$ is constructed by the output of the classifier $h_0$ for unlabeled data which is better estimation than maximum frequency that we used in first iteration. Then, these new set of feature vectors $X^{[1]}$ and labels $y^{[1]}$ are used to train a softmax classifier. We repeat this process iteratively for $T$ steps. At each intermediate step $t$, we have that $X^{[t]} = \{f(x_i^{[t-1]}, y_i^{[t]}) : x_i \in X\}$ is constructed by the output of the previous step for unlabeled cells. These features in fact induce a \emph{deep-learning-like} framework where at each stage, we are adding non-linear transformations from the previous stage. However, unlike deep models, we do not train the previous layers but those are fixed to the values we learned before. Due to space constraints, this discussion is only included with the appendix (Section \ref{sec:deep}). We finally output $h_T$. At each stage, we use a multi-class softmax as a classifier. Other choices for classification function are possible, namely multi-class logisitic regression \cite{kleinbaum2002logistic}, decision trees \cite{shalev2014understanding} etc. In this paper, we stuck with the choice of softmax regression as it gave good empirical performance as shown in the experiments sections. 

In some record fusion applications, it might not be possible to get a large set of labelled entities. In such situations, augmenting the training set with additional points might be very helpful. Even in cases where we have a large number  of training examples, data augmentation can prove to be helpful. We generate and add artificial entities from existing entities using standard transformation techniques. The output is a set of additional clusters that are legitimate to be used for training. Due to space constraints the details of data augmentation are included in the appendix. (Section \ref{sec:augment}). 

\vspace{-4pt}
\subsection{Inferencing with Variable Domain Size}
\label{section:inferenceAlgorithm}
\vspace{-7pt}
During inference, our goal is to predict the correct value for each column for each of the entities $E_k$. We have two different type of inference. For a column $j$, we have learned a classifier $f_j$ which outputs a $\rho_j$-dimensional vector for each cell. If the $E_k$ is part of $G_U$ then $y^{[T]}_U$ is the prediction. But if $E_k\not\in \mc C$, we first use $h_0, \ldots, h_T$ to obtain the feature vector of the cells in $E_k$. Then, for each of the $|E_k|$ values for the $j^{th}$ column, we can use our classifier to obtain $|E_k|$ different $\rho_j$-dimensional vectors. Now, we have two problems. The first is \textit{ how to fuse the $|E_k|$ predictions of the clusters into a single output?}. The second question is given a $\rho_j$-dimensional output \textit{how to do we use that to get one value for that cell?}

Both of these questions can be answered in several different ways. For the first question, our approach is to simply take an average of all the different vectors to obtain a single vector. Other more sophisticated approaches are possible but for now this gave us a good empirical performance. Again, for the second question, we choose the index which has the maximum value for the probability vector amongst the first $|E_k|$ indices and ignored the output values in indices from $|E_k|+1$ up to $\rho_j$ this analogous to re-normalize the distribution over the first $|E_k|$ indices and then take the maximum from that range. Again, instead of an $\argmax$ other probabilistic approaches like selecting an index with probability proportional to its value is possible. But the simple approach gave good empirical performance as shown by our extensive experiments. 

\vspace{-2pt}
\section{Experiments}
\label{sec:exp}
\vspace{-1pt}

We evaluate our record fusion framework using real datasets with various rules. We answer the following questions: (1) how well does record fusion framework work as data fusion system compared to the state-of-the-art data fusion systems when the source information of the entries is known.
\begin{table}
\vspace{-15pt}
	\center
	\scriptsize
	\caption{Datasets used in our experiments.}
	\label{tab:datasets}
	\begin{threeparttable}
		\begin{tabular}{|c|c|c|c|c|}
			\hline
			{\bf Dataset} & {\bf Size} & {\bf Clusters} & {\bf Attributes} & {\bf \# Sources}\\ \hline
			Flight & 57222 & 2313 & 6 & 37 \\
			Stock 1 & 113379 & 2066 & 10 & 55 \\
			Stock 2 & 107260 & 1954 & 8 & 55 \\
			Weather & 43003 & 13689 & 6 & 11 \\
			Address & 3287 & 494 & 6 & N/A\tnote{*} \\
			\hline
		\end{tabular}
		\begin{tablenotes}
			\item[*] N/A = Address dataset basically has been generated without sources.
		\end{tablenotes}
	\end{threeparttable}
	\vspace{-10pt}
\end{table}
(2) how well does it perform when the source of entries are not available compared to \cite{deng2017unsupervised} and Majority Vote. (3) what is the impact of different representation contexts on data fusion. (4) how well our cluster augmentation and iterative algorithm can solve the problem of learning from noisy and incomplete data. We use five benchmark datasets with different domain properties and usage described in Table~\ref{tab:datasets}. We compare our approach, referred to as $HF_S$ when we have sources and $HF_W$ when sources are unavailable, against several the-state-of-art methods. (see Appendix Section \ref{exp:setup} for more details).

\begin{table*}[]
\center
\scriptsize
\caption{Precision's Median, Average, and Variance of different methods for different datasets.}
\label{tab:endres}
\begin{threeparttable}
\begin{tabular}{c c|c c c c c c| c c c}
\shortstack{Dataset\\ ($\mathcal{T}$ size)}& Prec & $HF_S$ & Count & ACCU & CATD & SSTF & SlimFast & $HF_W$ & MV & USTL+MV \\ \hline \hline
\multirow{3}{*}{\shortstack{Flight\\ (5\%)}} & Med & \bf 0.998 & 0.901 & 0.878 & 0.952 & 0.739 & 0.220 & \bf 0.959 & 0.296 & 0.305\\
 & Avg &  0.998 & 0.882 & 0.892 & 0.939 & 0.732 & 0.241 & 0.947 & 0.176 & 0.337\\
 & StE & $3.4 \times10^{-4}$ & $0.0$ & 0.005 & 0.013 & 0.021 & 0.003 & 0.008 & 0.202 & 0.105\\ \hline
\multirow{3}{*}{\shortstack{Stock 1\\ (5\%)}} & Med & \bf 0.997 & 0.815 & 0.906 & 0.971 & 0.688 & 0.323 & \bf 0.985 & 0.051 & 0.050\\
 & Avg & 0.997 & 0.862 & 0.917 & 0.941 & 0.632 & 0.343 & 0.989 & 0.035 & 0.062\\
 & StE & 0.002 & 0.021 & 0.014 & 0.023 & 0.005 & 0.006 & 0.003 & 0.024 & 0.034\\\hline
\multirow{3}{*}{\shortstack{Stock 2\\ (5\%)}} & Med & \bf 0.988 & 0.840 & 0.853 & 0.823 & 0.779 & 0.767 & \bf 0.938 & 0.765 & 0.856\\
 & Avg & 0.991 & 0.825 & 0.812 & 0.737 &0.652 & 0.795 & 0.935 & 0.782 & 0.856\\
 & StE & 0.014 & 0.006 & 0.013 & 0.009 &0.031 & 0.028 & 0.013 & 0.064 & 0.107\\ \hline 
\multirow{3}{*}{\shortstack{Weather\\ (5\%)}} & Med & \bf 0.997 & 0.909 & 0.702 & 0.706 & 0.537 & 0.517 & \bf 0.794 & 0.721 & 0.738\\
 & Avg & 0.992 & 0.849 & 0.762 & 0.707 & 0.613 & 0.541 & 0.787 & 0.741& 0.732\\
 & StE & 0.004 & 0.018 & 0.009 & 0.017 & 0.011 & 0.003 & 0.021 & 0.060 & 0.023\\ \hline
 \multirow{3}{*}{\shortstack{Address\\ (10\%)}} & Med & n/a\tnote{*} & n/a & n/a & n/a & n/a & n/a & \bf 0.912 & 0.817 & 0.822\\
 & Avg & n/a & n/a & n/a & n/a & n/a & n/a & 0.899 & 0.740 & 0.780\\
 & StE & n/a & n/a & n/a & n/a & n/a & n/a & 0.019 & 0.231 &0.183\\ \hline \\
\end{tabular}
\vspace{-8pt}
\begin{tablenotes}
  \item[*] n/a = Address dataset basically has no source information, so source-needed algorithms cannot be executed.
  \end{tablenotes}
\end{threeparttable}
\vspace{-4pt}
\end{table*}

\vspace{-4pt}
\subsection{End-to-End Performance}
\label{exp:e2e}
\vspace{-7pt}
Table~\ref{tab:endres} summarizes the precision's Median, Average, and Variance of methods and as it shows, our method consistently outperforms all other methods. For {\it Flight},{\it Stock 1}, {\it Stock 2}, and {\it Weather}, we set the amount of training data to be $5\%$ of the total dataset. For Address, we set the percentage of training data to be $10\%$ (corresponding to 40 clusters) since Address is small.
 In the no sources information case, we see improvements of 70 points for {\it Flight} and {\it Stock 1}. More importantly, we find that our method is able to achieve low standard error in all datasets despite the different cluster and true record representation distribution in each dataset. This is something that seems challenging for prior data fusion methods and reduce their results consistency and reliability. Despite the fact that source information is an important factor for other algorithms, $HF_W$ can obtain high precision.
 This is because $HF_W$ models estimates the actual data distribution by extracting source signatures using attribute correlation from datasets. For instance, for {\it Address}, we see that \emph{MV} can find many of the true record representations---it has high precision---indicating that most true record representations correspond to statistical frequency. Overall, our method achieves an average precision of $91\%$ without sources information, and an average precision of $99\%$ when sources information available across these diverse datasets, while the performance of competing methods varies significantly and they are not consistent on all datasets.


\begin{figure*}
  \centering
  \makebox[\textwidth][c]
  {\includegraphics[width=1.1\textwidth]{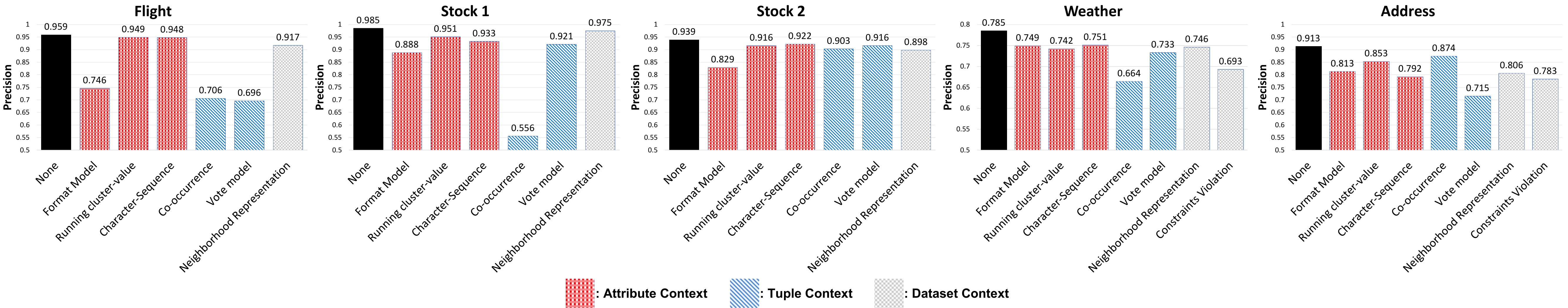}}
  \vspace{-10pt}
  \caption{Ablation studies to evaluate the effect of different representation models.}
  \label{fig:singAblation}
  \vspace{-10pt}
\end{figure*}

\vspace{-4pt}
\subsection{Representation Ablation Study}
\begin{figure}
\centering
\begin{minipage}{.4\textwidth}
  \centering
  \captionsetup{width=.9\linewidth}
  \includegraphics[width=\linewidth]{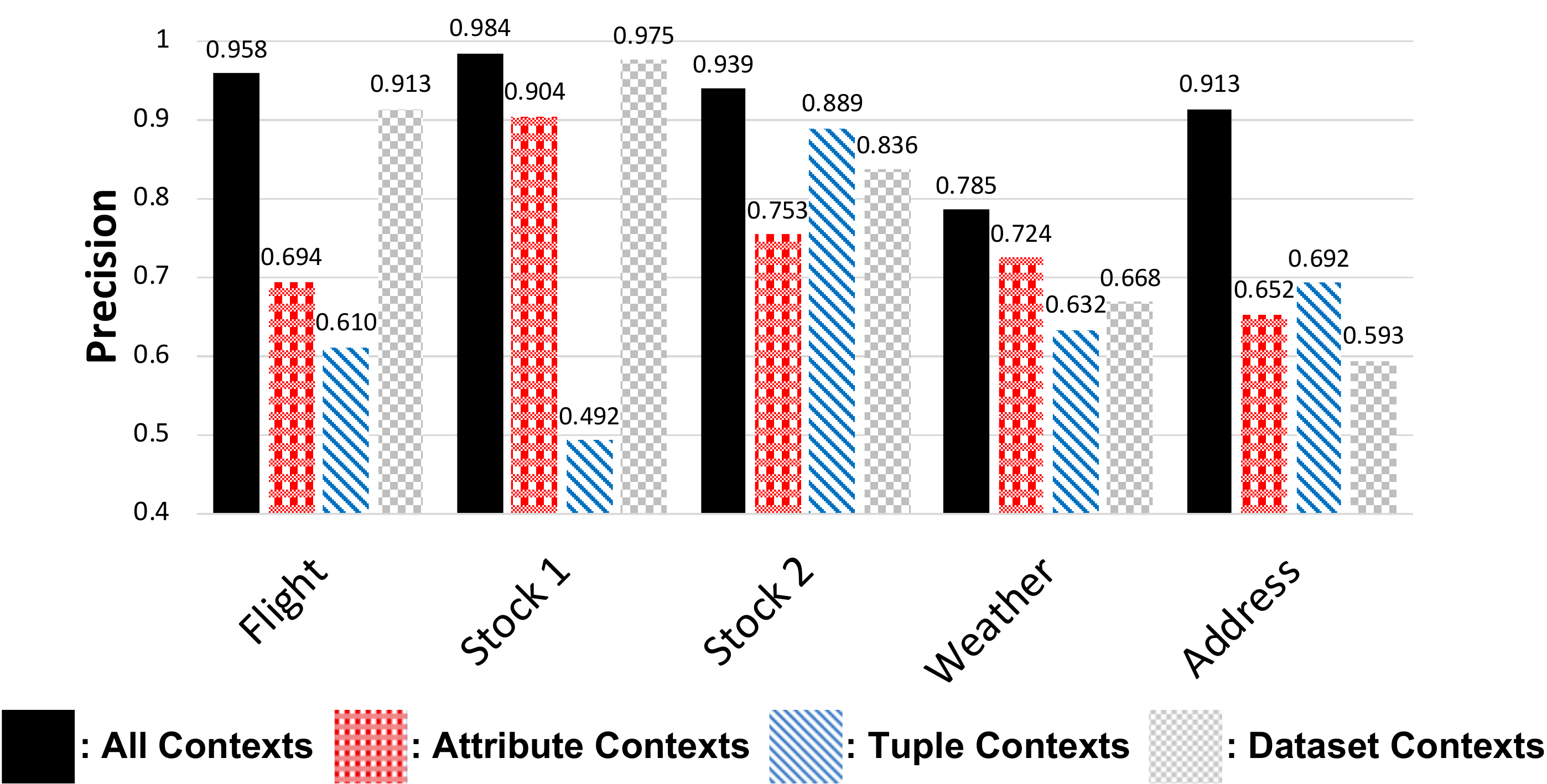}
  \captionof{figure}{Ablation studies to evaluate the effect of different representation model groups.}
  \label{fig:groupAblation}
\end{minipage}%
\begin{minipage}{.3\textwidth}
  \centering
  \captionsetup{width=.9\linewidth}
  \includegraphics[width=\linewidth]{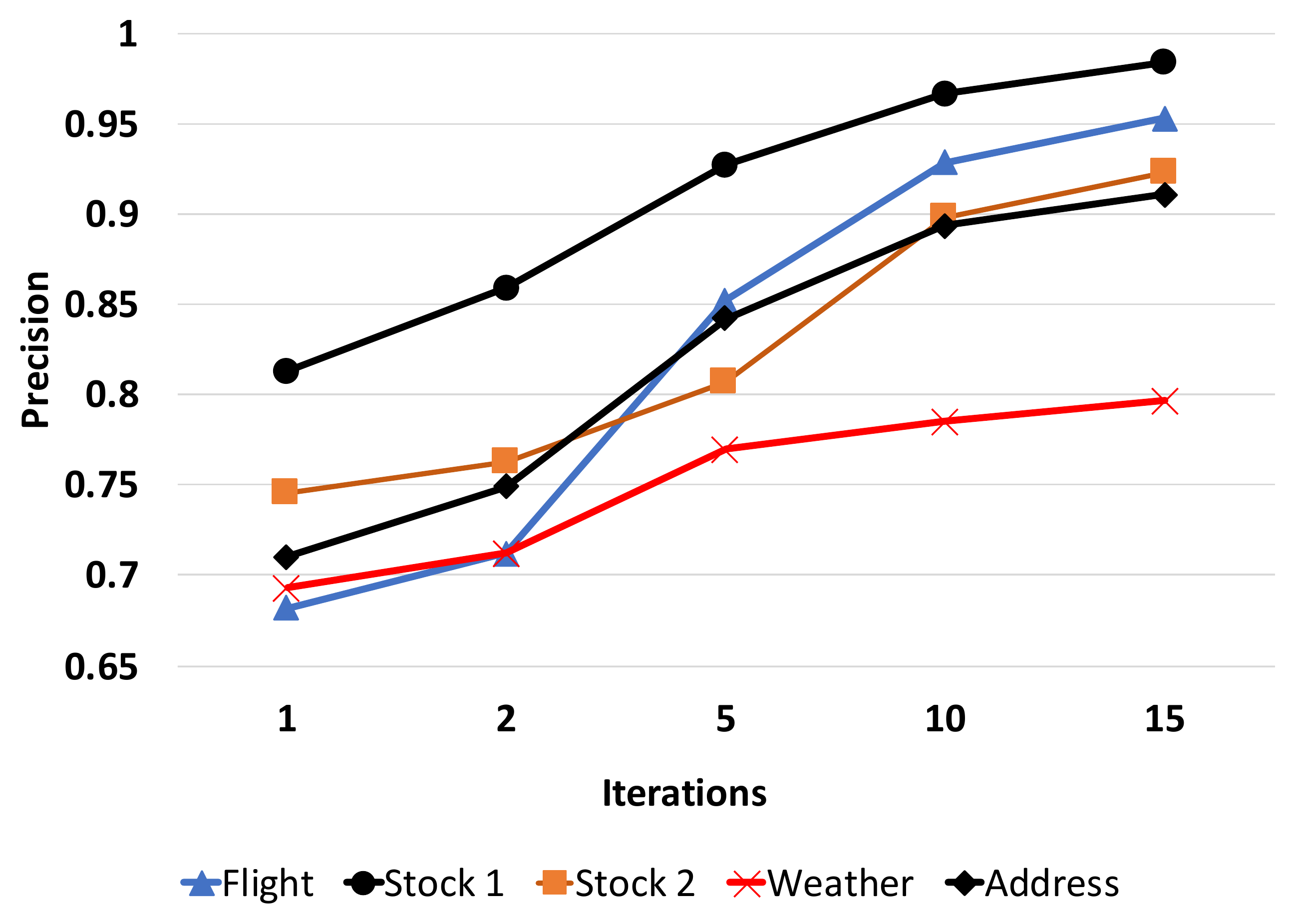}
  \captionof{figure}{The effect of increasing the number of clusters via data augmentation.}
  \label{fig:iter}
\end{minipage}%
\begin{minipage}{.3\textwidth}
  \centering
  \captionsetup{width=.9\linewidth}
  \includegraphics[width=\linewidth]{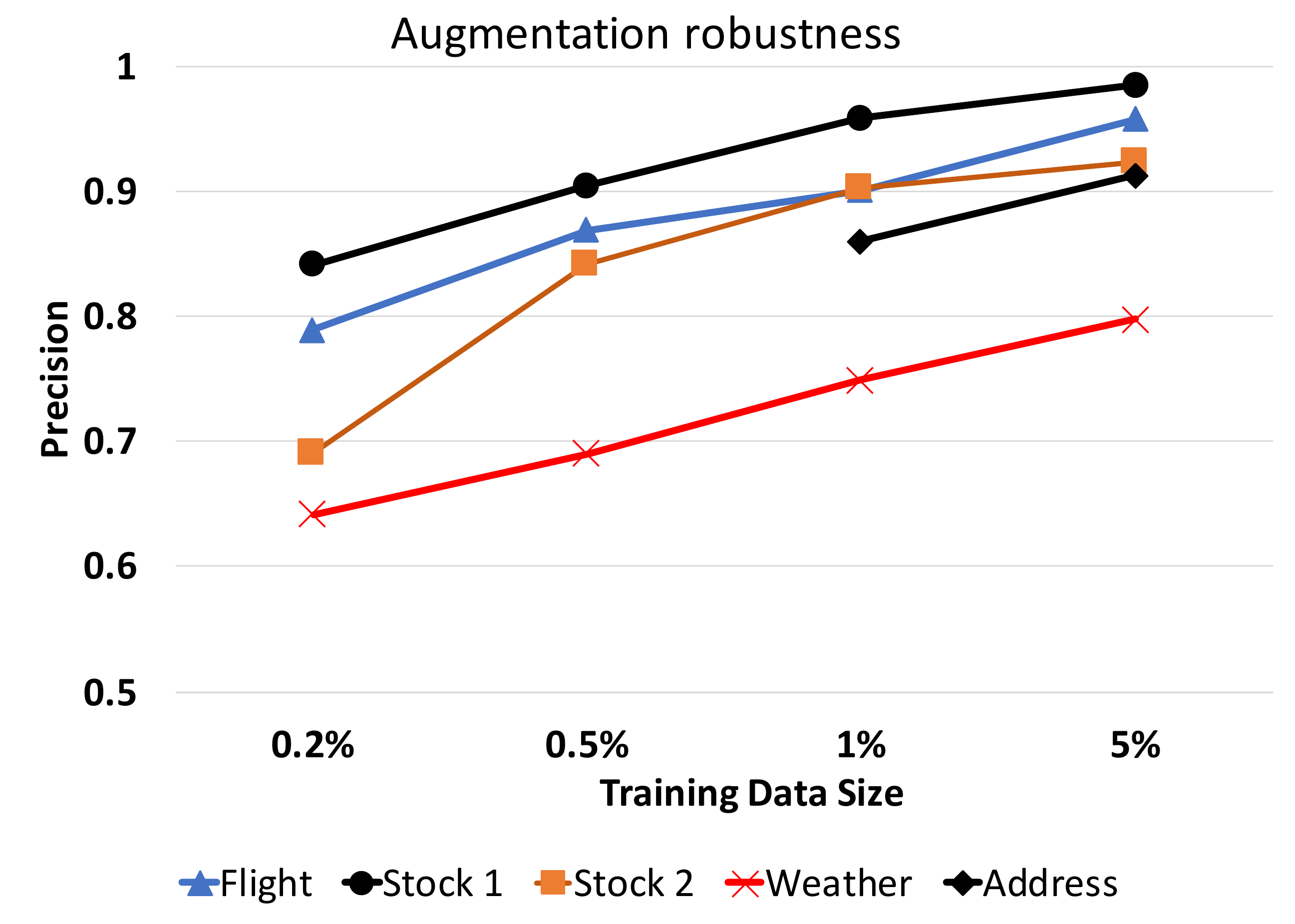}
  \captionof{figure}{The effect of increasing the number of clusters via data augmentation.}
  \label{fig:rob}
\end{minipage}
\vspace{-20pt}
\end{figure}

\vspace{-7pt}
We perform an ablation study to evaluate the effect of different representation models on the quality of our system. Specifically, we compare the performance of $HF_W$ when all representation models are used versus variants of $HF_W$, where a set of representation models is removed at a time.
\vspace{-10pt}
\paragraph{Single Representation Effect:} In Figure \ref{fig:singAblation}, we report the precision of the different variants as well as the original $HF_W$. It is shown that removing any feature has an impact on the quality of predictions of our model. More importantly, we find that different representation models have a different impact on various datasets. For instance, the most significant drop for {\it Stock1} and {\it Weather} is achieved when the co-occurrence model is removed, while for {\it Flight} and {\it Address}, the highest drop is achieved when the voting model is removed. 
Therefor, the representation models that we considered have a positive impact on the performance of our system. As it can be seen in Figure \ref{fig:singAblation}, for example, in dataset {\it Fight}, we see that removing \emph{Running-cluster value} representation from model has the minimum impact on the  performance, and if we see the parameters of this representation after it trained, they have values that almost ignore the impact of this signal.
\vspace{-10pt}
\paragraph{Group Contexts Effect}
Figure \ref{fig:groupAblation} shows the effect of a group of representation models corresponding to different contexts. Removing any contexts group has an impact on the quality of predictions of our model. Furthermore, for datasets that have various properties, different context groups have the most prominent effect on the performance of $HF_W$. This validates our design of considering representation models from different contexts. Therefore, it is necessary to leverage cluster representations that are informed by different contexts to provide robust and high-quality data fusion solutions.

\vspace{-4pt}
\subsection{Effects of Iterations on Performance}
\vspace{-7pt}

In this experiment, we validate the importance of the iterative process to improve learning performance. Figure \ref{fig:iter} shows the results of \emph{HF} for a various number of iterations. The results validate that as the number of iterations increases, we were able to get more accurate predictions. This observation has significant meaning for the performance of \emph{HF} as getting more accurate predictions in each iteration results in recalculating the dynamic features more accurately in each round. For instance, in Weather, \emph{HF} was able to achieve precision less than $0.7$ with only one iteration; however, after $15$ iterations, the precision was improved over $10$ points. Therefore, the recurrent process is an effective approach for calculating accurately dynamic features.

\vspace{-4pt}
\subsection{The Augmentation Process Robustness}
\vspace{-7pt}

Figure \ref{fig:rob} shows the performance of the augmentation process for different training data sizes. In summary, we find that the augmentation is robust and can achieve high precision even if the training data consist only $0.2\%$ of the clusters in the dataset.

\vspace{-5pt}
\section{Conclusion}
\label{sec:conc}
\vspace{-8pt}
We introduce a machine learning framework for {\it record fusion}, which the underlying challenge in two well-known classical problems: {\it data fusion} \cite{dong2009integrating,dong2013compact,rekatsinas2017slimfast} and {\it golden record} \cite{deng2017unsupervised}. We learn rich data representation models, and resolve the training data shortage via data augmentation.  We iteratively obtain and apply a model that can predict the correct label for unlabeled data. Our proposal outperformed previously proposed models, especially for the absence of source information.

\section{Broader Impact}
Record fusion is the main technology in almost all entity resolution and knowledge reconciliation efforts, which are crucial steps in building large scale knowledge bases and knowledge graphs. These consistent knowledge bases are powering many important downstream applications, including question answering and training large machine learning models. The arms race to construct these knowledge graphs among all major tech companies such as Google, Microsoft, and Amazon is a strong evidence on the impact of record fusion solutions. Most current record fusion methods are even simple methods to estimate the reliability of sources, or complicated and hard-to-maintain rule-based engines that do not scale with web-scale knowledge graphs. We believe that our proposal to automate record fusion, while taking into account all previous approaches as signals/features, provides a principled, holistic and an extensible solution that scales well to modern large knowledge graphs. We are currently in talks to deploy this proposal with major knowledge graph creators.

\vspace{1em}
{\fontsize{15}{100}\selectfont \textbf{Appendix}}
\begin{appendix}
\section{Related Work}
\label{sec:rel}
Several pieces of research have been done on combining data from multiple sources. Bleiholder et al. \cite{bleiholder2006conflict} surveyed existing strategies for resolving inconsistencies in structured data. The data fusion methods can be categorized into four main regimes:

\begin{itemize}
    \item \textbf{Na\"ive method}: In this method, all sources have a vote, and the correct value for each object is decided by choosing the value that has maximum votes among all the conflicting values.
    \item \textbf{Source-based}: The main goal of these methods is to calculate how accurate each source is. More specifically, the votes of the sources do not have the same "weight." The importance of each vote depends on the quality of the source.
    \item \textbf{Relation-based} These methods use the main idea of Source-based methods. They also consider the correlation between the sources (e.g., if a pair of sources copy from each other).
    \item \textbf{Transformation-based} These methods reduce cluster size by transforming values to each other and use human-in-the-loop to fuse remained set.
\end{itemize}

In the field of discovering dependencies between data sources, many works have been done as well. In \cite{dong2009integrating}, Dong et al. applied Bayesian analysis to decide on dependencies between sources. In \cite{dong2010global}, Dong et al. also consider various types of copying on different data items. Moreover, the authors in \cite{dong2009truth} explore the idea of integrating data and determining the way the sources are interacting with each other by examining the update history of the sources.

Besides, there has been a lot of research in the field of evaluating trustworthiness resulting in algorithms such as {\it PageRank} which assigns trust based on link analysis and TrustFinder\cite{yin2008truth} which decides about the importance of a source based on its behavior in a P2P network. Moreover, in \cite{dong2012less}, Dong et al. examine the problem of selecting a subset of sources before integration. The authors claim that by choosing only the sources that can be beneficial for their algorithm, they can achieve higher performance than by using all the available sources and data. 

Furthermore, in \cite{pasternack2010knowing}, Pasternack and Roth solve the data fusion problem by creating an iterative model. The main idea of their fact-finding algorithm was to incorporate prior knowledge from the users into their process, to integrate data from conflicting claims. 
In \cite{rekatsinas2017slimfast}, Rekatsinas et al. propose the SLiMFast framework to solve the data fusion problem as a learning and inference problem over discriminative probabilistic graphical models. The method of this paper is also the first that came with guarantees on its error rate for the estimation of the source accuracy.

Finally, in \cite{deng2017unsupervised} human-in-the loop  is used to solve entity consolidation; instead of using sources, the system simulates source information for entity resolution by asking information from an oracle. To the best of our knowledge, this is the only method that can work without source information.

\subsection{Deduplication process}
Since the input of record fusion problem is the output cluster of deduplication process, we give the following example.
\begin{example}
Figure \ref{fig:example2} shows a tabular data containing tuple id's, person names, occupation, and address and illustrates the task of a typical deduplication approach. Deduplication produces a table with the clustering of those records, where each cluster refers to the same real-world entities.  The process of finding the true records for entities is called \say{golden record problem}. In this setting, there is no information about sources, as all records might have come from the same source.

\begin{figure}[h]
  \centering
  \includegraphics[width=\columnwidth]{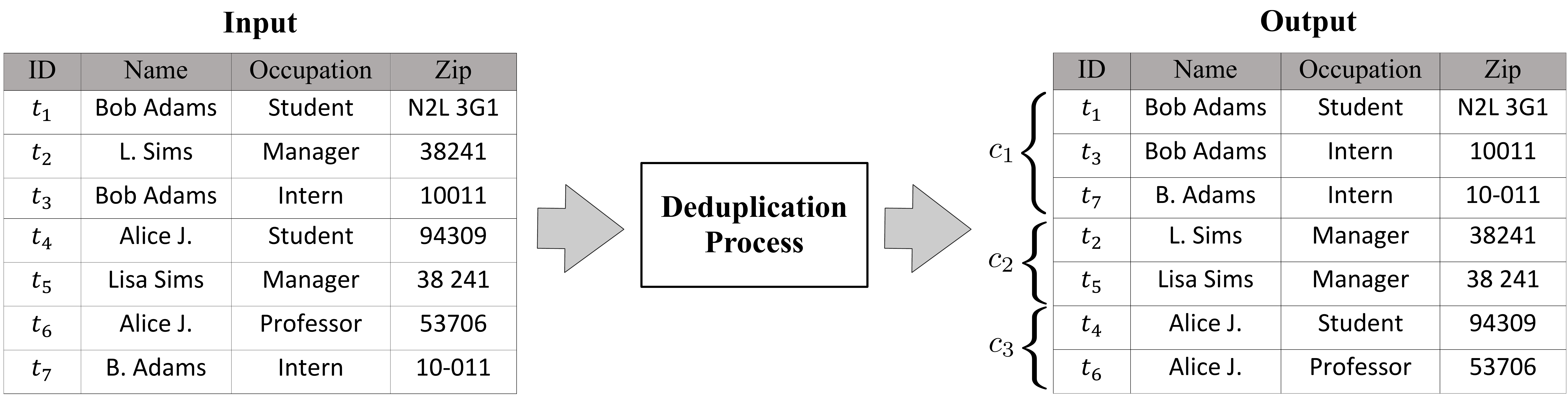}
  \caption{ A typical deduplication task.}
  \label{fig:example2}
\end{figure}

\end{example}

\section{Feature design}
\label{sec:features}
Regarding to the feature design described in Section \ref{sec:featuredesign}, we have Alg \ref{alg:attributeFeatures} for Attribute-Level features \ref{sec:attrfeat}, Alg \ref{alg:recordFeatures} for Record-Level features \ref{sec:recfeat}, and Alg \ref{alg:databaseFeatures} for Database-Level features \ref{sec:datafeat}.

\begin{algorithm}
	\SetAlgoLined
	\KwIn{Database $D$.\\
		\hspace{35pt}Cell $d_{ij}$ with row $i$ and column $j$ such that $Id(i) = k$.\\
		\hspace{33pt} Set $E_{kj}$ of  all possible values for the $k^{th}$ entity on the $j^{th}$ column.\\
		\hspace{33pt} An embedding matrix $M$ which maps all elements in $E_{kj}$ to a vector in $\mb R^m$.\\
	}
	\vspace{5pt}\KwOut{Vector $v_{ij}$ representing the attribute-level features for $d_{ij}$.}
	
	\vspace{10pt} Map $d_{ij}$ is a string $t_{ij}$ over the alphabet $\{A, N, S\}$\;
	
	\ForEach{character $c$ in $d_{ij}$}{ 
		If $c$ is a letter then map  $c$ to `A' \;
		If $c$ is a number  then map  $c$ to `N' \;
		If $c$ is a space character then map  $c$ to `S' \;
	}
	
	Let $u_{ij}$ be the nine dimensional vector representing the counts of all $2$-grams of $p_{ij}$\;
	
	\ForEach{$d'_{ij}$ in $E_{kj}$}{ 
		Compute the $m$-dimensional embedding of $d'_{ij}$ using the matrix $M$. Denote it by $a'_{ij}$  \;
	}
	
	Compute the average of the embeddings and denote by $a$.\;
	
	Let $w_{ij} = dist(a_{ij}, a)$\;
	
	Make $\rho_j$-dimensional vector $x_{ij}$ of zeros\;
	
	\If{$\neg$(first iteration)}{
		Use the model prediction in previous iteration and put one in corresponding dimension in $x_{ij}$\;
	}
	
	\vspace{5pt}\textbf{return} $v_{ij} := [u_{ij},  w_{ij}, x_{ij}]$
	\caption{Attribute-level features}
	\label{alg:attributeFeatures}
\end{algorithm}

\begin{algorithm}[h]
	\SetAlgoLined
	\KwIn{Database $D$.\\
		\hspace{36pt}Cell $d_{ij}$ with row $i$ and column $j$ such that $Id(i) = k$.\\
		\hspace{33pt} Set $E_k$. All rows corresponding to the $k^{th}$ entity. \\
		
	}
	\vspace{5pt}\KwOut{Vector $v_{ij}$ representing the record-based features for $d_{ij}$.}
	
	\vspace{10pt} \ForEach{column $j' \neq  j$}{ 
		Let $n_{j'}$ be the number of occurrences of $(d_{ij}, d_{ij'})$  over $E_k$\;
		Let $m_{j'}$ be the number of occurrences of $d_{ij'}$  over attribute $j'$ in $E_k$\;
	}
	
	Let $u_{ij} = [\ldots, \frac{n_{j'}}{m_{j'}}, \ldots]$  be the $c-1$ dimensional vector representing the co-occurrence counts of the given attribute $d_{ij}$\;
	
	Let $w_{ij} = \frac{t}{|E_k|}$where $t$ is the number of occurrences of $d_{ij}$  over $E_{kj}$ \;
	
	\vspace{5pt}\textbf{return} $v_{ij} := [u_{ij},  w_{ij}]$
	\caption{Record-level features}
	\label{alg:recordFeatures}
\end{algorithm}

\begin{algorithm}[h]
	\SetAlgoLined
	\KwIn{Database $D$.\\
		\hspace{36pt}Source matrix $S$ of size $n \times k$. (Optional)\\
		\hspace{36pt}Cell $d_{ij}$ with row $i$ and column $j$ such that $Id(i) = k$.\\
		\hspace{33pt} Set $E_k$ of all rows belonging to the $k^{th}$ entity. \\
		\hspace{33pt} Set $\Sigma_j$ of denial constraints for the $j^{th}$ column.\\
		\hspace{33pt} An embedding matrix $M$ which maps all elements of  the $j^{th}$ column to $\mb R^m$.\\
		\hspace{33pt} An embedding matrix $Q$ which maps all rows to $\mb R^q$.
	}
	\vspace{5pt}\KwOut{Vector $v_{ij}$ representing the database-level features for $d_{ij}$.}
	
	\vspace{10pt} \ForEach{row $d_{i'} \in E_k$}{ 
		Compute the $q$-dimensional embedding for $d_{i'}$ using matrix $Q$. Call it $a_{i'}$\;
		Compute the $m$-dimensional embedding for $d_{i'j}$using matrix $M$. Call it $b_{i'j}$\;
		Let $n_{i'} = [a_{i'}, b_{i'j}]$
	}
	
	Let $n = avg(n_{i'})$  be the average embedding vector \;
	Let $u_{ij} = dist(n_i, n)$\;
	
	\vspace{5pt} Let $w_{ij} = []$ \\
	\ForEach{$\sigma \in \Sigma_j$}{
		Compute  $x = |\{ \text{rows make violation with row i w.r.t. }\sigma\}|$. That is compute the number of violations of the constraint $\sigma$ assuming the value $d_{ij}$ is correct\;
		$w_{ij}.append( x)$
	}
	Let $S_i$ be the $k$-dimensional vector indicating the source information for the $i^{th}$ row.\\
	
	\vspace{5pt}\textbf{return} $v_{ij} := [u_{ij},  w_{ij}, S_i]$
	\caption{Dataset-level features}
	\label{alg:databaseFeatures}
\end{algorithm}

\section{Learning algorithm}
\subsection{Relation to iterative learning}
Another way to view Alg. \ref{alg:il} is through the framework of iterative learning or `dynamic' features. Consider the original set of features in the data $X$. To this set of features, we have a set of dynamic vectors. Denote the new dataset by $X'$. If $X_S$, the static part of $X$ has dimension $\nu$ and the dynamic parts of $X$ has $\psi$ then $X$ and $X'$ have dimension $\nu + \psi$.  The $\psi$-dimensional vector represents the set of dynamic features. We initialize the dynamic features by the all zero vector. Observe that Alg. \ref{alg:il} is identical to the algorithm with dynamic features as stated in the corollary below. 

\begin{algorithm}
	\small
	\SetAlgoLined
	\KwIn{
		Training dataset $Z' = (X', y)$. \\
		\hspace{35pt} Number of iterations $T$.
	}
	\KwOut{Classifier $h: X' \rightarrow y'$}
	
	\vspace{10pt} Let the dynamic features of $X$ be equal to zero.\\
	\vspace{5pt}\For{$t=1$ to $T$}{
		Let $h_t$ be the softmax classifier obtained by training on $(X', y)$\\
		Let $y'$ be the set of new labels, that keep the training set label and update the rest\\ 
		Let the new features of $X'$ be equal to $f(X',y')$. \\
	}	
	\vspace{5pt}\textbf{return} $h_T$
	\caption{Iterative learning with dynamic features}
	\label{alg:ilAlternate}
\end{algorithm}

\begin{corollary}
	Given $Z = (X, y)$ such that $y \in \{1, \ldots, \rho\}$. Let $X' = (X_S, \mb 0_{\psi})$ where  $\mb 0_{\psi}$ denotes the $\psi$-dimensional vector of all zeros. Consider an iterative version of softmax classification on the set $Z' = (X', y)$ defined in Alg. \ref{alg:ilAlternate}. Then, the Alg. \ref{alg:il}  is identical to the formulation in Alg. \ref{alg:ilAlternate} with dynamic features. 
\end{corollary}
\subsection{Relation to deep learning}
\label{sec:deep}
In the previous sections, we saw how our algorithm can be viewed as a stagewise model or a model with dynamic set of features. In this section, we look at another interpretation; namely, its relation to deep learning. At each stage, our model does the following computation.
\begin{align*}
& y^{[t]}_U=\{y^{[t]}_i=h^{[t-1]}(X^{[t-1]}):\forall i\in I_U\}\\
& y^{[t]}=y_T\cup y^{[t]}_U	\\
& X^{[t]} \enspace=\enspace f(X^{[t-1]}, y^{[t]}) \enspace \text{where we have that }\\
&h_{t-1}(X^{[t-1]}) \enspace =\enspace  softmax(\thinspace  W_{t-1} \enspace X^{[t-1]} \thinspace)
\end{align*}

Note that in standard deep learning architectures, $X^{[t]} = \zeta_{t-1}(X^{[t-1]})$. In our architecture, we also concatenate it with the original set of features $X_S$. While in deep learning the number of hidden states is varied and is a hyper-parameter, in this framework the number of `hidden states' is fixed at $\psi + \nu$ (the sum of dimension of dynamic and static vectors).

Another important distinction is regarding the training algorithm. The standard deep networks are trained with backpropogation which updates the weights of all the layers of the network in one backward pass. In this case, we train the network in a greedy manner. We first train the first layer of the network by using the softmax loss on its output. The learned weights are then used to compute the input features of the next layer. And then the process is repeated. Observe that this corresponds to `freezing' the weights of the previous layers and only training the weights of the current layer. We refer to this way of training as \textit{greedy-layerwise} training.

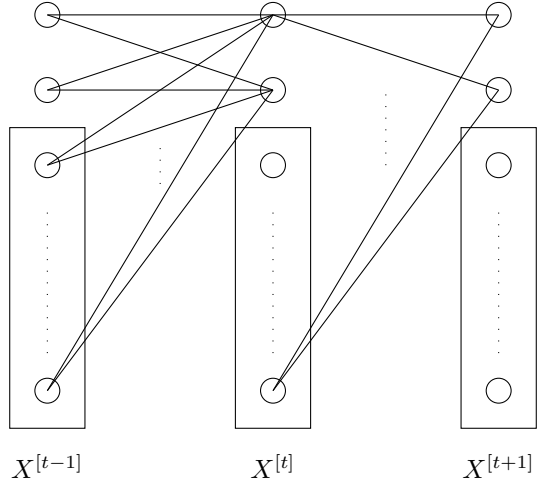
\begin{figure}
	\centering
	\begin{tikzpicture}
	\draw (0.5,1) node[circle,draw]{};
	\draw [loosely dotted](0.5,1.5) node{} --  (0.5,3.5) node{};
	\draw (0.5,4) node[circle,draw]{};
	\draw (0.5,5) node[circle,draw]{};
	\draw (0.5,6) node[circle,draw]{};
	\draw (0,0.5) node{} -- (0,4.5) node{}-- (1,4.5) node{} -- (1,0.5) node{} --cycle;
	\draw (0.5,0) node{$X^{[t-1]}$};
	
	\draw (3.5,1) node[circle,draw]{};
	\draw [loosely dotted](3.5,1.5) node{} --  (3.5,3.5) node{};
	\draw (3.5,4) node[circle,draw]{};
	\draw (3.5,5) node[circle,draw]{};
	\draw (3.5,6) node[circle,draw]{};
	\draw (3,0.5) node{} -- (3,4.5) node{}-- (4,4.5) node{} -- (4,0.5) node{} --cycle;
	\draw (3.5,0) node{$X^{[t]}$};
	
	\draw (6.5,1) node[circle,draw]{};
	\draw [loosely dotted](6.5,1.5) node{} --  (6.5,3.5) node{};
	\draw (6.5,4) node[circle,draw]{};
	\draw (6.5,5) node[circle,draw]{};
	\draw (6.5,6) node[circle,draw]{};
	\draw (6,0.5) node{} -- (6,4.5) node{}-- (7,4.5) node{} -- (7,0.5) node{} --cycle;
	\draw (6.5,0) node{$X^{[t+1]}$};

	\draw (0.5,1) node{} --  (3.5, 5) node{};
	\draw (0.5,1) node{} --  (3.5, 6) node{};
	\draw (0.5,6) node{} --  (3.5, 5) node{};
	\draw (0.5,6) node{} --  (3.5, 6) node{};
	\draw (0.5,5) node{} --  (3.5, 5) node{};
	\draw (0.5,5) node{} --  (3.5, 6) node{};
	\draw (0.5,4) node{} --  (3.5, 5) node{};
	\draw (0.5,4) node{} --  (3.5, 6) node{};
	
	\draw (3.5,1) node{} --  (6.5, 5) node{};
	\draw (3.5,1) node{} --  (6.5, 6) node{};
	\draw (3.5,6) node{} --  (6.5, 5) node{};
	\draw (3.5,6) node{} --  (6.5, 6) node{};
	
	\draw [loosely dotted](2,3.75) node{} --  (2,4.25) node{};
	\draw [loosely dotted](5,4) node{} --  (5,5) node{};	
	\end{tikzpicture}
	
	\caption{ The nodes in the box represent the original set of features $X$. The two nodes outside represent the `dynamic features'. At every layer $t$, non-linear function of the features of the previous layer are added to the model. Compared to traditional deep learning, in this case the original set of input features are always passed to next layer.}
	\label{fig:deep}
	\vspace{-20pt}
\end{figure}

The logical steps are illustrated in Fig. \ref{fig:deep}. The input features are shown in the box and are copied through all the layers of the network. The nodes outside the boxes correspond to the  `dynamic'  features which have a dimension of $\psi$. Each layer first computes a linear mapping of the features of the previous layer using the computation $W X^{[t-1]}$. Then a non-linearity (in this case a softmax function) is applied to the linear map. Hence, deeper and deeper layers represent more and more complex non-linear transformations of the original input space.

\section{Data Augmentation}
\label{sec:augment}
In the previous sections, we described our featurization techniques and the learning algorithm. Together they are sufficient as a learning framework for the record fusion problem. However, in this section we go a step further and also propose a data augmentation mechanism.

In some record fusion applications, it might not be possible to get a large set of labelled entities. In such situations, augmenting the training set with additional points might be very helpful. Even in cases where we have a large number  of training examples, data augmentation can prove to be helpful in the following way. Recall that the domain of record fusion suffers from the homogeneity versus heterogeneity problem. That is, all the entities are quite `different' from one another while records within the same entity (or cluster) are `similar'. In such cases, a data augmentation approach introduces clusters which are similar to existing clusters and enables better model generalization.

\begin{algorithm} 
	\SetAlgoLined
	\KwIn{
		Database $D$.
	}
	\KwOut{Augmented entities $\hat E_1, \ldots, \hat  E_o$}
	
	\vspace{5pt}Select a `source entity' $E_s \in D$ uniformly at random. \\
	\ForEach{cell $d_{ij} \in E_s$ which does not belong to the ground truth}{ 
		Use Alg. \ref{alg:stringToAlphabet} to map $d_{ij}$ to another string $g_{ij}$ over the alphabet $\Lambda$.\\
		Select a `target entity' $E_t \in D$ and then select row $d_{i'} \in E_t$ uniformly at random.\\
		Using the same procedure as above, map $d_{i'j}$ to $g_{i'j}$.\\ 
		Compute $b$, the longest common sub-string between $g_{ij}$ and $g_{i'j}$. \\
		Use the reverse mapping (Alg. \ref{alg:stringToAlphabet}) to map $b$ back to the augmented string $\hat d_{ij}$. \\
		Add $\hat d_{ij}$ to the augmented entity $\hat E$
	}
	Repeat the above procedure to get $o$ augmented entities.
	\caption{Entity augmentation}
	\label{alg:data_aug}
\end{algorithm}

Our data augmentation procedure is described in Alg. \ref{alg:data_aug}. Before we discuss the procedure in greater detail, lets first introduce some notation.

\begin{definition}[Format alphabet]
	Let $\mc S_1$  be the set  of all the letters of the English alphabets (small case and capitalized). Let $\mc S_2 = \{s \in S_1^+ :  |s| > 1\}$ be the set of all strings of letters of length greater than one. Similarly, let $\mc T_1 = \{0, \ldots, 9\}$ and $\mc T_2 = \{s \in T_1^+ : |s| > 1\}$. Also denote by $\mc U = \{space, \#, \$, ? , \ldots\}$ the set of `special' characters. Define the set of symbols $$\Lambda = \{\mc S_1, \mc S_2, \mc T_1, \mc T_2\} \bigcup_{s \in \mc U} s$$
\end{definition}

Given a string in our database, we map it to the set of symbols $\Lambda$. Roughly, this captures the `format' of the input string. For example, consider that the cell of a database has the value \textit{`New York-\#401H3'}. Our mapping algorithm will map it to the string \textit{$\mc S_2 space \mc S_2 \# \mc T_2 \mc S_1\mc T_1$}.  Inuitively, this captures the format that the input contains a letters followed by a space followed by letters etc. In this way we represent the source string $d$ as a format string $g$. Next, we repeat the same procedure to get the target format string $g'$ from the target string $d'$. This gives us information that the source string could also have the format $g'$ instead of $g$. Hence, our augmentation procedure involves `editing' the string $d$ to obtain another string $\hat d$ such that the format of $\hat d$ is the same as $g'$. Thus by repeating this process for all the cells of the entity, we obtain the augmented entity $\hat E$. 

Two details are missing from the discussion in the above paragraph. Firstly, how the mapping algorithm works and secondly how to `edit' a given string with format $g$ to match another format $g'$. Lets look at the former first. The editing or augmentation step then follows from that.

\begin{algorithm} 
	\SetAlgoLined
	\KwIn{ String $d$.}
	\KwOut{
		String $g \in \Lambda^+$ which is mapping onto the format space.\\
		\hspace{38pt} $\tau^{-1}$ which maps each character of $g$ to a substring of $d$.  
	}
	
	\vspace{5pt} Let $g = \emptyset$\\
	\ForEach{character $c \in d$}{
		If $c$ is an english alphabet letter, add $\mc S_1$ to $g$.\\
		Else if $c$ is a number, add $\mc T_1$ to $g$.\\
		Else add $c$ to $g$.
	}
	Change $g$ by replacing all consecutive occurences of $\mc S_1$ of length greater than one by $\mc S_2$. \\
	Similarly, replace all consecutive occurences of $\mc T_1$ of length greater than one by $\mc T_2$. \\
	
	\ForEach{character $f \in g$}{
		Let $\tau^{-1} (f)$ be the substring in $d$ which mapped to $f$. 
	}
	
	Return $g$ and $\tau^{-1}$.
	\caption{Format mapping}
	\label{alg:stringToAlphabet}
\end{algorithm}

The algorithm works by first mapping the given string to another string over the alphabet $\{\mc S_1, \mc U_1\} \cup_{s \in \mc U} s$. We then map all consecutive occurences of $\mc S_1$ to $\mc S_2$. For example, let the source string be $d = $\textit{`New York-\#401H3'}. Then, $d$  is first mapped to \textit{$\mc S_1\mc S_1\mc S_1space\mc S_1\mc S_1\mc S_1\mc S_1-\#\mc T_1\mc T_1\mc T_1\mc S_1\mc T_1$}.  In the second step, all the consecutive occurences of $\mc S_1$ and $\mc T_1$ are mapped to $\mc S_2$ and $\mc T_2$ respectively. Hence, the final format representation for $d$ is $g = \mc S_2 space \mc S_2-\#\mc T_2 \mc S_1 \mc T_1$. The mapping $\tau^{-1}$ keeps track that the first $\mc S_2$ corresponds to the string \textit{`New'}, the seconf $\mc S_2$ corresponds to  \textit{`York'} and so on.

Once the mapping algorithm is known, the `editing' or augmentation process is fairly straightforward. Let the target string be $d' = $\textit{`Toronto-\#21LG'} which maps to the format $g' = \mc S_2-\#\mc T_2 \mc S_2$. In this case the longest common sub-string between $g$ and $g'$ is $\mc S_2-\#\mc T_2$. Using the (inverse) mapping $\tau^{-1}$, this gives back the augmented string as $\hat d = $\textit{York-\#401}. And this is added as a cell to the corresponding augmented entity.

\section{Experiments}
\subsection{Experimental Setup}
\label{exp:setup}
We describe the datasets, metrics, and settings that we use in our experiments. We use five benchmark datasets with different domain properties and usage described in Table~\ref{tab:datasets}.

\begin{table*}
\center
\small
\caption{ Data augmentation performance for various amounts of training data $\mathcal{T}$.}
\label{tab:tdaug}

\begin{tabular}{c l|c c c c c c c}{}
Dataset & $\mathcal{T}$ & $HF_{w/o AUG}$& 0.05 & 0.1& 0.3&0.5&0.7&1\\ \hline \hline
\multirow{3}{*}{Flight} & 5\% & 0.779 & 0.869 & 0.919&\bf 0.958 & 0.949 & 0.946 & 0.928\\
 & 10\% & 0.802 & 0.893 & 0.924&\bf 0.967& 0.956 & 0.950 & 0.937\\ \hline
\multirow{3}{*}{Stock 1} & 5\% & 0.826 & 0.942 & \bf 0.985&0.928 & 0.939 & 0.920 & 0.914\\
 & 10\% &0.843 & 0.966 &\bf 0.992&0.957 & 0.942 & 0.934 & 0.944\\ \hline
 \multirow{3}{*}{Stock 2} & 5\% & 0.825&\bf 0.923 & 0.913&0.902 & 0.903 & 0.903 & 0.901\\
 & 10\% &0.853 &\bf 0.938 & 0.928&0.924 & 0.923 & 0.915 & 0.913\\ \hline
  \multirow{3}{*}{Weather} & 5\% & 0.737 & 0.749 & \bf 0.798&0.763 & 0.755 & 0.749 &0.760\\
 & 10\% & 0.770 & 0.782 &\bf 0.805&0.790 & 0.774 & 0.766 & 0.763\\ \hline
  \multirow{3}{*}{Address} & 5\% & 0.837 & 0.874 & 0.904&0.903 & \bf 0.913 &  0.912 & 0.904\\
 & 10\% & 0.869 & 0.904 & 0.915&0.914 &\bf 0.930 & 0.927 & 0.910\\ \hline
\end{tabular}
\vspace{-10pt}
\end{table*}
\textit{Stock 1 and 2} contain data from $55$ stock sources from popular financial aggregators such as \emph{Yahoo! Finance}, \emph{Google Finance}, and \emph{MSN Money}, official stock-market websites such as \emph{NASDAQ}, and financial-news websites such as \emph{Bloomberg} and \emph{MarketWatch}. \emph{Stock 1} contains 2066 objects (clusters) and the ground truth is created by  assuming that \emph{NASDAQ} always provides the correct value. \emph{Stock 2} contains 1954 objects and the ground truth is created by taking the majority value provided by five stock data providers \cite{li2012truth}.

\textit{Flight} is a benchmark dataset that contains $37$ sources from the flight domain. The sources include $3$ airline websites (\emph{AA}, \emph{UA}, Continental), $8$ airport websites (such as \emph{SFO}, \emph{DEN}), and $26$ third-party websites, including \emph{Orbitz}, \emph{Travelocity}, etc. The dataset focused on $2313$ flights departing from or arriving at the hub airports of the three airlines (\emph{AA}, \emph{UA}, and Continental). Each cluster is a specific flight on a particular day \cite{li2012truth}. The ground truth was created by taking the majority value of three sources, including the source \emph{AA}, which always has the correct value.

\textit{Weather} is collected for $30$ major USA cities from $11$ websites about every $45$ minutes. We consider \emph{(city, time)} as the key. There are in total $33$ collections in a day, thus the dataset contains $990$ clusters. The attributes are manually mapped, and there are $6$ distinct attributes. 
The ground truth is created by taking the majority value provided by all the sources.

\textit{Address} reflects applications for discretionary funding to be allocated by the New York City Council. For each record, we select attributes that represent legal information, address and geographical properties of location. The minimum size of each cluster is two and the ground truth has been extracted from \emph{ISBNsearch} organization website. An interesting feature of the \emph{Address} dataset is that it does not contain any source information.

These datasets are used as standard benchmarks for data fusion algorithms. Notice that we are given data rules (denial constraints) only for \textit{Weather} and \textit{Address} datasets. (see Figure \ref{fig:rob})

We compare our approach, referred to as $HF_S$ when we have sources and $HF_W$ when sources are unavailable, against several data fusion methods. First, we consider five baseline data fusion models that they need sources information: \textit{Counts}: This corresponds to Na\"ive Bayes. Source accuracies are estimated as the fraction of times a source provides the correct value for an object in ground truth. \textit{ACCU}: This is the Bayesian data fusion method introduced by Dong et al. \cite{dong2009integrating} (without source copying). \textit{CATD}: A fusion method introduced by Li et al. \cite{li2014confidence} and extends source reliability scores with confidence intervals to account for sparsity in source observations. \textit{SSTF}: This data fusion method by Yin et al. \cite{yin2011semi} leverages semi-supervised graph learning to exploit the presence of ground truth data. \textit{SlimFast}:  A data fusion framework by Rekatsinas et al. \cite{rekatsinas2017slimfast} based on statistical learning over discriminative probabilistic models. 

We also compare to two approaches that require no sources information. \textit{Majority Vote (MV)}: In each cluster-attribute, we consider the maximum frequency value as the true record representation. \textit{USTL+MV}: This entity consolidation method, which was introduced by Dong et al. \cite{deng2017unsupervised}, uses human-in-the-loop to request user to verify the equivalence of records, and minimizes the number of queries by transforming values in an unsupervised way. Then the \emph{Majority Vote} can be used to obtain correct records. 

\begin{table*}[t]
\center
\small
\caption{ Iterative algorithm performance for various amounts of training data $\mathcal{T}$.}
\label{tab:tditer}
\begin{tabular}{c l|c c c c c}{}
Dataset/Aug & $\mathcal{T}$ & 1& 2 & 5& 10&15\\ \hline \hline
\multirow{3}{*}{Flight/0.3} & 5\% & 0.682 & 0.712 & 0.853& 0.930 & \bf 0.953\\
 & 10\% & 0.707 & 0.725 &0.881& 0.944& \bf 0.966\\ \hline
\multirow{3}{*}{Stock 1/0.5} & 5\% & 0.813 & 0.859 &  0.928&0.966 &\bf 0.984\\
 & 10\% &0.834 & 0.865& 0.933&0.969 &\bf 0.991\\ \hline
 \multirow{3}{*}{Stock 2/0.05} & 5\% & 0.746& 0.763 & 0.808&0.899 &\bf 0.923\\
 & 10\% &0.764 & 0.784 & 0.843&0.911 &\bf 0.937\\ \hline
  \multirow{3}{*}{Weather/0.1} & 5\% & 0.694 & 0.713 &  0.770&0.785 &\bf 0.797\\
 & 10\% & 0.714 & 0.742 & 0.784&0.799 &\bf 0.807 \\ \hline
  \multirow{3}{*}{Address/0.5} & 5\% & 0.710 & 0.749 & 0.843&0.896 &\bf 0.912\\
 & 10\% & 0.721 & 0.754 & 0.888&0.907 &\bf 0.929\\ \hline
\end{tabular}
\vspace{-15pt}
\end{table*}
\vspace{3pt}\noindent{\bf Evaluation Setup:} To measure precision, we use Precision (P) defined as the fraction of  true record representation predictions that are correct. For training, we split the available ground truth into three disjoint sets: (1)~a training set $T$, used to find model parameters; (2)~a validation set, which is used for hyper parameter tuning;  and (3)~a test set, which is used for evaluation. To evaluate different dataset splits, {\em we perform $50$ runs with different random seeds for each experiment}. To ensure that we maintain Precision, we report the median performance. The mean performance along with standard error measurements are also reported. {\em Seeds are sampled at the beginning of each experiment, and hence, a different set of random seeds can be used for different experiments.} We use {\it ADAM}~\cite{kingma2014adam} as the optimization algorithm for all learning-based model and train all models for 500 epochs with a batch-size of ten examples. We run Platt Scaling for 50 epochs. All experiments were executed on a 12-core Intel(R) Xeon(R) CPU E5-2603 v3 @ 1.60GHz with 64GB of RAM running Ubuntu 14.04.3 LTS.
\begin{figure}[ht]
  \centering
  \vspace{-5pt}
  \includegraphics[width=0.5\linewidth]{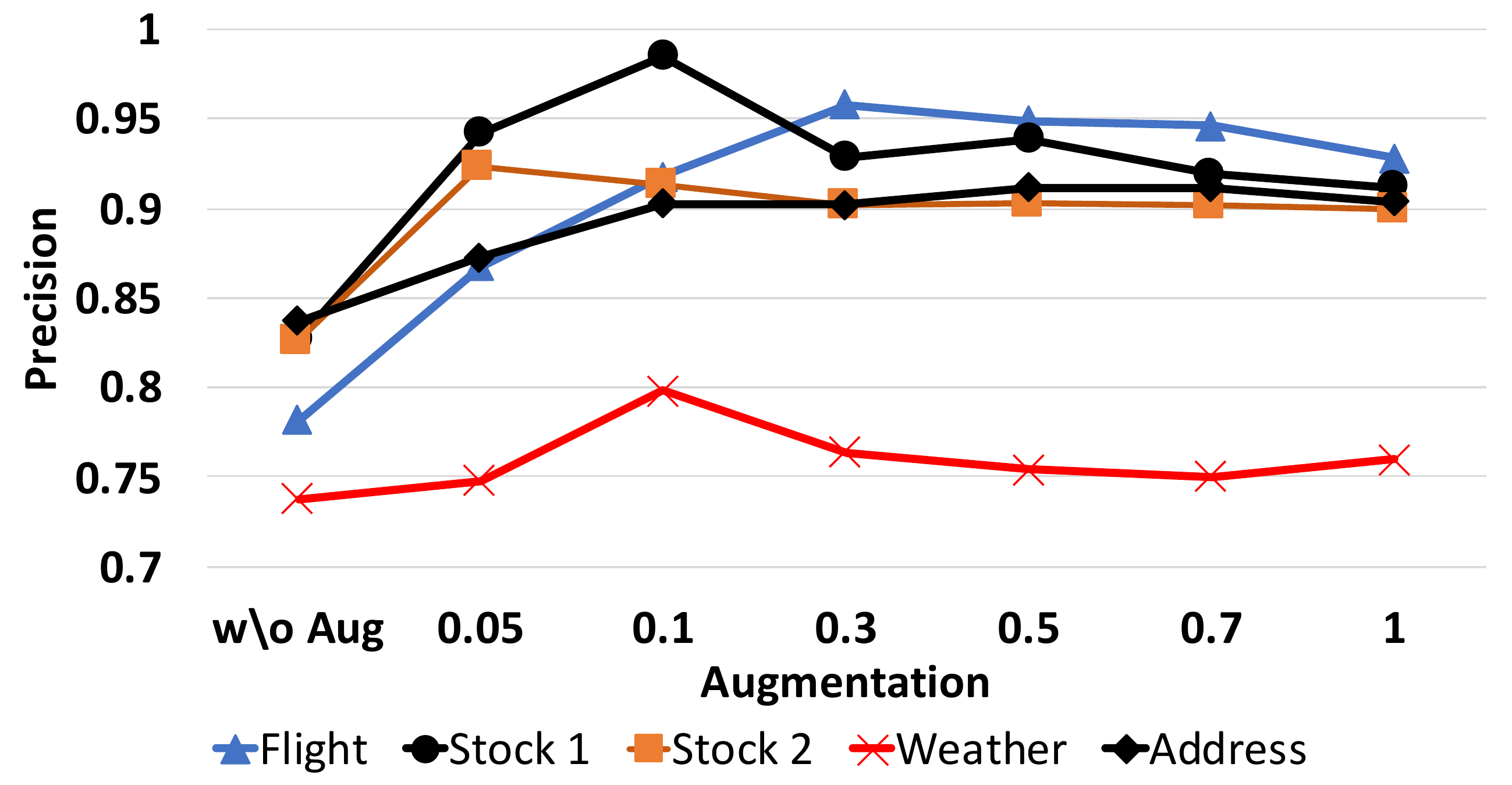}
  \caption{The effect of increasing the number of clusters via data augmentation.}
  \vspace{-5pt}
  \label{fig:aug}
\end{figure}

\subsection{Labeled Data Size Effect}
In these experiments, we evaluate the effect of the size of the training data on the performance of the augmentation policy and the iterative algorithm.
\subsubsection{Effect on Augmentation Performance}
Table \ref{tab:tdaug} shows the data augmentation performance for various amounts of training data. In all datasets, increasing the size of training data from $5\%$ to $10\%$ increase the performance. Moreover, in most datasets, the size of the labeled data does not affect the best augmentation ratio. Only in the {\it Address}, we see that to achieve the best precision, the augmentation ratio is $0.7$ when we have $5\%$ of the dataset as training data in contrast with the $0.5$ ratios when the training data are the $10\%$ of the dataset.  This is likely behavior is due to the fact that {\it Address} is generally small, thus by using only $5\%$ as training data, our augmentation algorithm needs to create more augmented clusters for training.

\subsubsection{Effect on Iterative Algorithm Performance}

In Table \ref{tab:tditer}, the iterative algorithm performance for various amounts of training data can be observed. As was expected, the increase in training data enhances the performance of the iterative algorithm. It can also be observed that \emph{HF} needs at least 15 iterations in order to converge to the best precision independently of the training size.

\subsection{Effects of Augmentation on Performance}
We evaluate the effectiveness of data augmentation to counteract the lack of training data. Figure \ref{fig:aug} shows that using data augmentation yields high-quality record fusion models for datasets with varying sizes and properties (as they were described in section \ref{exp:setup}). Hence, data augmentation is robust to different domains of properties.

We also evaluate the effect of excessive data augmentation: We manually set the ratio between the initial clusters and the lately generated cluster in the final training examples and use augmentation to materialize this ratio. Our results are reported in Figure \ref{fig:aug}. We see that peak performance is achieved when the ratio between the two types of clusters is about 10\% to 30\% for all datasets.We can conclude that data augmentation is an effective and robust way to counteract the lack of enough training data.

\section{Record Fusion System Overview}

\begin{center}
\begin{figure*}
  \centering
  \makebox[\textwidth][c]{\includegraphics[width=1.12\textwidth]{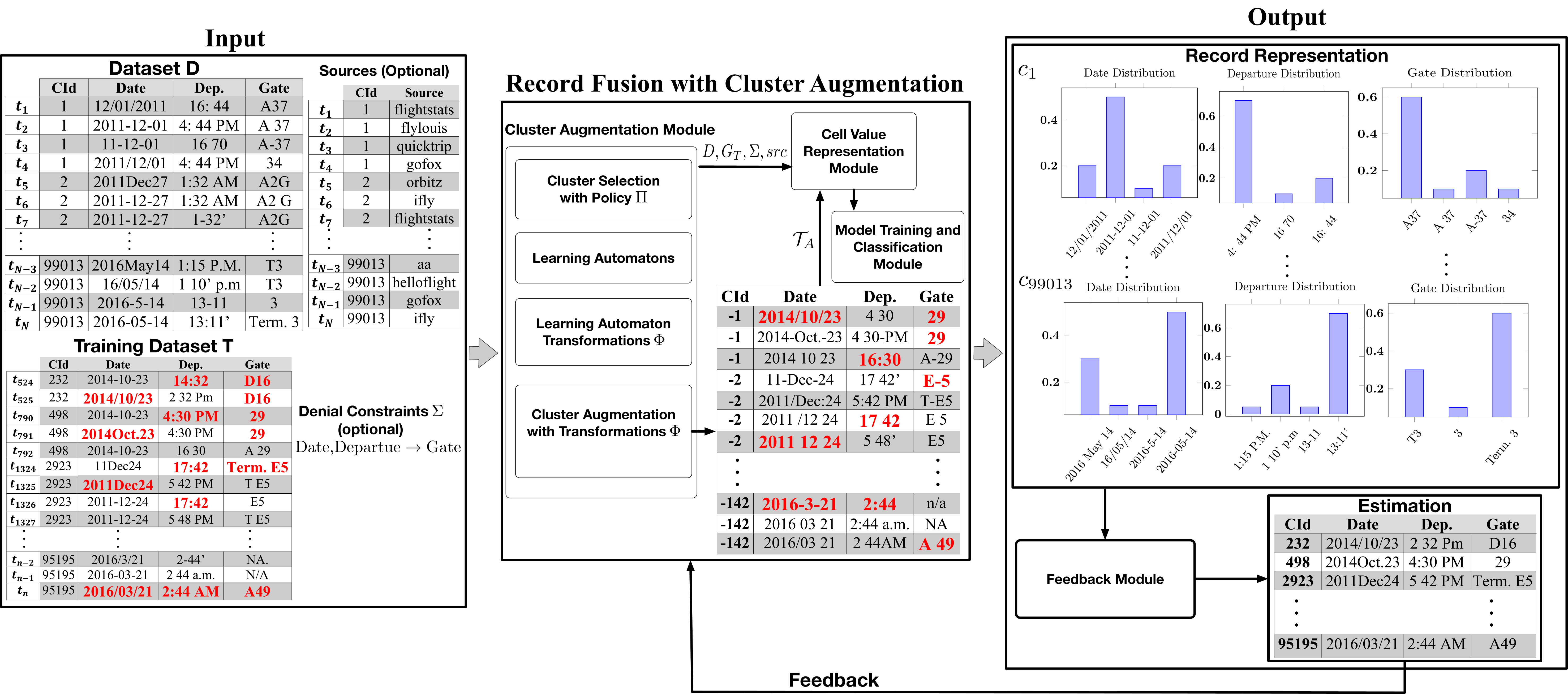}}
  \caption{Overview of Record Fusion with Augmentation.}
  \label{fig:overview}
\end{figure*}
\end{center}

An overview of how the different modules are connected is shown in Figure \ref{fig:overview}. First, Module 1 augments training data with additional artificial clusters. Then, Module 2 grounds the representation model of our record fusion model. Subsequently, the representation model is connected with the multi-class classifier model in Module 3, after generating record representation, the model gets feedback from Module 4, and so it changes the representation and the predictions. 

\end{appendix}

\bibliographystyle{abbrv}

\bibliography{recordfusion} 

\end{document}